\pdfoutput=1

\documentclass[11pt]{article}
\usepackage{arabtex}
\def\endabstract{\egroup}
\newcommand\tab[1][1cm]{\hspace*{#1}}
\usepackage[]{acl}
\usepackage{amssymb}
\usepackage{graphicx}
\usepackage{times}
\usepackage{latexsym}
\usepackage{utf8}
\usepackage{amsmath}
\usepackage{multirow}

\usepackage[T1]{fontenc}

\usepackage[utf8]{inputenc}

\usepackage[most]{tcolorbox}

\tcbset{textmarker/.style={%
        enhanced,
        parbox=false,boxrule=0mm,boxsep=0mm,arc=0mm,
        outer arc=0mm,left=6mm,right=3mm,top=7pt,bottom=7pt,
        toptitle=1mm,bottomtitle=1mm,oversize}}

\newtcolorbox{hintBox}{textmarker,
    borderline west={4pt}{10pt}{yellow},
    colback=yellow!10!white}
\newtcolorbox{importantBox}{textmarker,
    borderline west={6pt}{0pt}{red},
    colback=red!10!white}
\newtcolorbox{noteBox}{textmarker,
    borderline west={6pt}{0pt}{green},
    colback=green!10!white}

\newcommand{\warning}[1]{\begin{hintBox} #1 \end{hintBox}}

\usepackage{microtype}

\title{
\begin{center}
    \includegraphics[width=2.0cm]{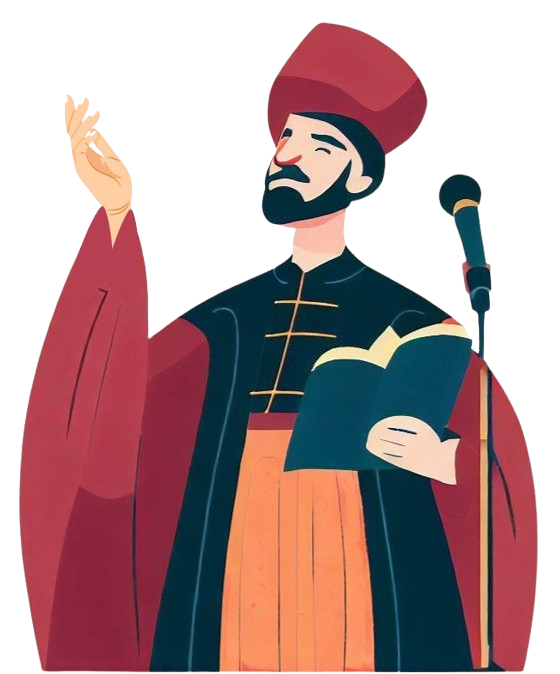}
\end{center}
Ashaar: Automatic Analysis and Generation of Arabic Poetry Using Deep Learning Approaches
\vspace{1cm}
}

\author{Zaid Alyafeai \and Maged S. Al-Shaibani \and Moataz Ahmed \\
King Fahd University of Petroleum and Minerals \\
Dhahran, Saudi Arabia \\
\texttt{g201080740@kfupm.edu.sa}
}

\begin{document}
\setcode{utf8}

\maketitle
\begin{abstract}
Poetry holds immense significance within the cultural and traditional fabric of any nation. It serves as a vehicle for poets to articulate their emotions, preserve customs, and convey the essence of their culture. Arabic poetry is no exception, having played a cherished role in the heritage of the Arabic community throughout history and maintaining its relevance in the present era. Typically, comprehending Arabic poetry necessitates the expertise of a linguist who can analyze its content and assess its quality. This paper presents the introduction of a framework called \textit{Ashaar} \footnote{\url{https://github.com/ARBML/Ashaar}}, which encompasses a collection of datasets and pre-trained models designed specifically for the analysis and generation of Arabic poetry. The pipeline established within our proposed approach encompasses various aspects of poetry, such as meter, theme, and era classification. It also incorporates automatic poetry diacritization, enabling more intricate analyses like automated extraction of the \textit{Arudi} style. Additionally, we explore the feasibility of generating conditional poetry through the pre-training of a character-based GPT model. Furthermore, as part of this endeavor, we provide three datasets: one for poetry generation, another for diacritization, and a third for Arudi-style prediction. These datasets aim to facilitate research and development in the field of Arabic poetry by enabling researchers and enthusiasts to delve into the nuances of this rich literary tradition.
\end{abstract}

\section{Introduction}
\label{sec:intro}
In a general setting, Arabic poetry could be divided into two forms: rhymed or measured and prose. Rhymed poetry was first introduced and theorized by \texttt{al-Farahidi} (711 – 786 A. D.) who categorized every poem into one of 15 different classes, later extended to 16, called meters or \textit{Buhur} as pronounced in Arabic. These meters govern how each poem should be constructed with specific rules called \textit{Arud} or \textit{Arudi Style}. The main constructs of Arud could be represented using \textit{Tafeelat} as plural or \textit{Tafeelah} as singular for easier memorization. Such constructs could be used to define how to create each meter using a finite set of rules. Another important part of Arabic poetry is \textit{Qafiyah} which refers to the end rhyme pattern or the rhyme scheme used in the poem. 
The construction of meters depends on diacritics which are special symbols assigned to each letter in the poem. These diacritics are categorized as either harakah or sukun.  Analyzing poems usually needs expertise in the field to figure out the consistent meter and find out issues if there are any. Poets, nevertheless, have an intrinsic ability to construct poems from a specific meter without the need to consult experts. Recently, in the modern era, many poets were influenced by western culture resulting in a new form of poetry called prose poetry. Prose poetry is loose in terms of rules but has some structure and rhythm although not in a strict format. Modern poets used poetry as a medium to express various emotions and feelings. 
Prose poetry is similar to English poetry in the way it is constructed but, due to its long history, Arabic poetry is richer in terms of metaphors and symbolism. 
\begin{figure*}[ht]
    \centering
    \includegraphics[width=1.0\textwidth]{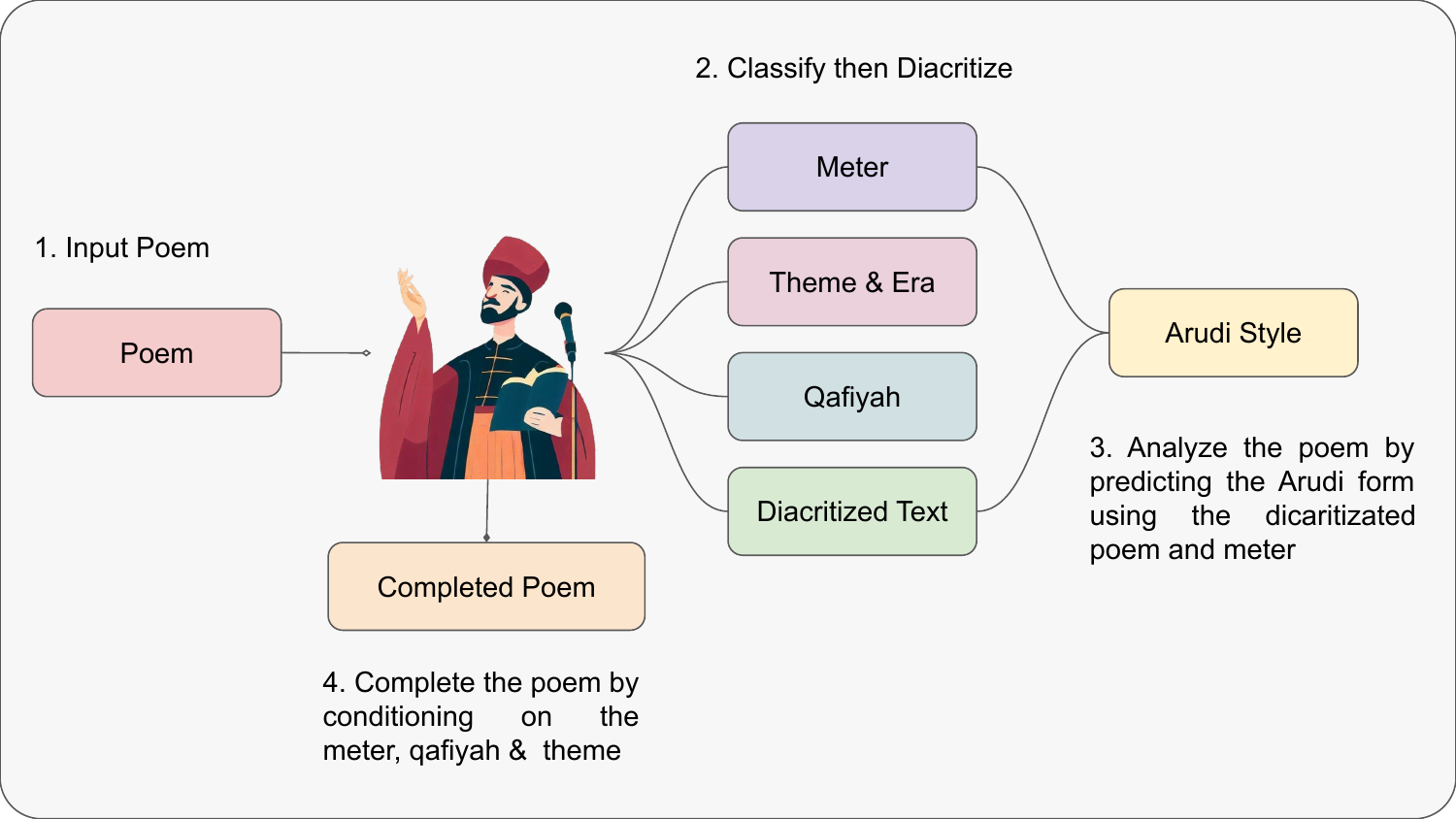}
    \caption{General pipeline for Ashaar.}
    \label{fig:pipeline}
\end{figure*}

In this paper, we utilize deep learning approaches to analyze and generate poetry. A high-level pipeline is shown in Figure \ref{fig:pipeline}. We summarize our contributions as the following:

\begin{enumerate}
    \item We create four public datasets: \textit{Ashaar dataset}\footnote{\url{https://huggingface.co/datasets/arbml/Ashaar\_dataset}} is a labeled dataset with meter, theme, era, etc. that could be used for conditional poetry generation. \textit{Ashaar diacritized}\footnote{\url{https://huggingface.co/datasets/arbml/Ashaar\_diacritized}} is a cleaned dataset with diacritized poems. \textit{Ashaar arudi}\footnote{\url{https://huggingface.co/datasets/arbml/Ashaar\_ardui}} is a dataset that gives gold Arudi representations for a given set of verses. \textit{Ashaar tafeelah}\footnote{\url{https://huggingface.co/datasets/arbml/Ashaar\_tafeelah}} which contains all the possible tafeelat for a given meter. 
    \item We provide five pre-trained models. Three classification models for era, theme, and meter. One pre-trained model for diacritization. And, a pre-trained model for conditional poetry generation. 
    \item We introduce a framework named \textit{Ashaar} for poetry analysis and generation. The analysis part uses the meter and diacritization models to predict the Arudi form. While, the generation part uses the meter, qafiyah, and theme to generate poem completion. 
\end{enumerate}

\section{Literature Review}
\label{sec:lr}
Many studies have been proposed to analyze and study the Arabic poetry metric system. Most of such efforts are directed towards linguistic libraries. \cite{ziyovuddinova2021arud} \cite{manna2021metrics}, \cite{paoli2001meters}, and \cite{maling1973theory} are just examples of the literary work advocated to the subject.

Below is a list of the tasks we found in the literature that deals with Arabic poetry from various aspects. These tasks include Authorship Attribution, meter classification, emotion and era classification, poetry identification from textual sources, poetry generation, and other miscellaneous tasks.

\subsection{Authorship Attribution}
\label{subsec:authorship_attribution}

In Arabic literature, there are many studies that dealt with authorship attribution in general text.  \cite{al2020ensemble}, \cite{altakrori2018arabic}, \cite{altheneyan2014naive}, and \cite{hajja2019authorship} are instances of various methods used to approach this problem for general Arabic text. For a special format of Arabic text like poetry, limited work has been proposed. 

\cite{ahmed2016authorship} used machine learning methods such as Support Vector Machines (SVM) and Sequential Minimal Optimization (SMO) to study the problem of the Authorship Attribution of Arabic poetry. The features they extracted from poetry cover characters, lexical syntactic, and semantic features. They applied their methods to a corpus of poems belonging to 54 poets. They achieved 98\% precision for SMO as the best score.

\cite{al2015authorship} attempted to approach this problem using Markov chains. They conducted their experiments on characters and other syntactically crafted features. The experiments were conducted on a dataset of 33 samples from 33 authors for training and another different 33 unknown samples for testing. They achieved more than 96\% accuracy score on the test set.

\cite{albaddawi2021pattern} developed a deep-learning model to identify poetry authors. The features they used are a fusion of the character embeddings and an LSTM-based pre-trained meter classification model. This architecture was evaluated on a dataset of more than 100k verses from 10 famous Arabic poets. They achieved around 81\% accuracy.

On a different direction, \cite{omer2017arud} utilized Arud words encoding as binary features for prose Authorship Attribution. They compare this set of features to another baseline of only considering the most frequent 100 words. They showed that their method is superior compared to this baseline. They tested their method on two different sets of Arabic and English texts.

\subsection{Meter Classification}

The work on textual Arabic meter classification can be divided into two main categories based on the techniques used. The first category covers the techniques that are rule-based while the second category approached this problem using deep learning methods. The prominent drawback of the first approach is that it requires the poetry text to be diacritized either fully or partially. Another characteristic of this category is that it has been evaluated on relatively small datasets as compared to the second category. The largest evaluation study is reported by \citep{alabbas2014basrah} consisting of less than 7k verses. Below, we survey the literature for both approaches. 

\textbf{Traditional Machine Learning} Several methods have been proposed to classify Arabic poetry meters. \cite{mohammad2009naive} proposed a Naive Bayes while \cite{ismail2010expert} proposed a knowledge-based framework. \cite{saleh2012arabic} filed a patent for a system that classifies poetry from acoustic as well as textual input. \cite{alnagdawi2013finding}, \cite{alabbas2014basrah}, and  \cite{abuata2018rule} proposed rule-based systems. \cite{ahmed2017program} introduced a rule-based system to analyze the rhyme of the poem. \cite{berkani2020pattern} created a matching pattern approach where the verse is matched against a curated set of meter patterns. \cite{zeyada2020proposed} suggested a system that extends the meter classification task for modern Arabic poetry, albeit that modern Arabic poetry does not need to follow a meter, unlike classical poetry. \cite{alqasemi2021arabic} evaluated traditional machine learning techniques on a partial dataset proposed by \cite{al2020meter}.

\textbf{Deep Learning} \cite{yousef2019learning} is the first work that utilizes deep learning for this task for all the 16-meter classes. They also tried to approach this task in English and Arabic languages. It is worth noting that Arabic poetry classes are 16 as compared to only 4 meters in English. This makes the task more complex to approach for Arabic. In their research, they introduced, APCD, a large dataset of 1.5M verses of Arabic poetry. The model they proposed is RNN-based. The results they achieved are 96.38\% and 82.31\% for Arabic and English respectively. \cite{al2020meter} proposed a GRU-based model to classify Arabic poetry meters. The model is a 5-layer stacked GRU followed by two dense layers. The dataset introduced in this research is MetRec \cite{al2020metrec} constituting more than 55.4K verses of 14-meter classes. The result they achieved is 94.3\% on the accuracy score. \cite{abandah2020classifying} extended the work done by \cite{al2020meter} and \cite{yousef2019learning} on this task. They introduced a larger RNN-based model evaluated on a dataset of poetry and prose, 17 classes in total. They introduced the APCD2 dataset which is an extended version of APCD with the prose class. In their research, they mark the use of diacritics as optional in contrast to \cite{al2020meter} where these characters are removed from the input stream. The results they reported crossed 97\% accuracy on this task.

\subsection{Emotion and Era classification}
\label{subsec:emotion_era_classificaiton}
In the literature, there is a lot of focus to work on era classification as compared to theme classification. 

\textbf{Theme Classification} \cite{alsharif2013emotion} investigated the promise of machine learning methods to address the task of Arabic poetry emotion classification. The dataset they collected consists of 1,231 Arabic poems variable in length with four major emotional classes: Retha (Elegiac), Ghazal (love or romance), Fakhr (pride or honor), and Heja (Satire). They experimented with Naive Bayes classification, SVM, Voting Features Intervals (VFI), and Hyperpipes. They reported the results of their experiments in precision, recall, and F-measure. They showed that VFI outperforms others in terms of F-Measure with a result of around 73\%.

\textbf{Era Classification} Depending on a set of literary features, Arabic scholars divided the Arabic poetry timeline into a couple of time segments based on either political status or literary features specific to that period of time or location. These segments are called eras. \cite{abbas2019classification} tried to classify Arabic poetry into its recitation era. The era classes they worked on are 5 ranging from \textit{pre-Islamic} era to \textit{Andalusian} era. The dataset comprised a set of more than 30k poems belonging to these different classes. Various machine learning methods have been experimented with this dataset. They showed that Multinomial Naive Bayes achieved the best performance with an F1-score of 68.8\% and a kappa score that is very close to 0.4. \cite{orabi2020classical} proposed a deep learning-based approach to address era classification. The dataset they used is scraped from the web. It consists of 60,377 poems in classical Arabic recited by 739 poets. They developed two deep learning-based models and compared their performance. The first is a classification model with FastText \citep{bojanowski2017enriching} embeddings while the second is a CNN-based model. They showed that the CNN model was superior achieving more than 91\% result on F1-score in terms of binary classification into modern and non-modern poetry. \cite{khorsheed2013comparative} proposed a comprehensive study on Arabic text classification with different textual styles including poetry. The poetry dataset they used comprised 1.95K documents with different 6 classes. They tried different features selections methods along with different machine learning classifiers. The best classification results they achieved were for the C5.0 classifier with 80\% on average for all styles and 50\% accuracy for poetry only. They attributed this low results to the difficulty of the classification task on creative materials like poetry. \cite{gharbat2019discovering} evaluated various classification models for classify poetry from \textit{Abbasid} and \textit{Andalusian} eras. The evaluated models are logistic regression, random forest, decision trees, and SVM. They evaluated these models on a curated dataset from the web. The dataset contains around 10,895 hemistiches (half a verse) of 15 random poems by 15 poets. Their experiments showed that SVM achieved the best performance.

\subsection{Poetry Generation}
With the recent advancement of deep learning approaches, there were many attempts in the literature to generate Arabic poetry. 

\cite{talafha2019arabic} proposed a GRU-based approach to generate Arabic poetry. They trained their model on a dataset comprising more than 20.1k poems with 80.5K verses collected from the web. For evaluation, they conducted two types of evaluations: quantitative and qualitative. For quantitative analysis, the BLEU score was used. For qualitative, they involved human subjects to evaluate the generated poetry.


\cite{beheitt2022automatic} proposed a GPT-based model to generate poetry. The model was trained from scratch. The methodology they followed is first training the GPT model on a newswire dataset to develop language understanding and then fine-tuning the model on a poetry dataset. The model was evaluated on BLEU as well as human evaluation. They showed that their approach outperformed other approaches that are based on elementary architectures like RNNs and GRUs.
\cite{elkaref2022generating} evaluated the poetry generation task on two transformer-based models with two different promoting settings. The evaluated models are BERTShared \cite{rothe2020leveraging} and GPT-j \cite{gpt-j} and the prompting methods are rhythm or topic based. The dataset used for this research is a fused collection of an earlier version of Ashaar and a public dataset published in GitHub \cite{AhmedAbd87:online}. They found out that GPT-J is better at capturing the rhyme while BERTShared is better at generating fluent poems.
\cite{abboushi2023toward} fine-tuned AraGPT2 \cite{antoun2020aragpt2} to generate poems. The dataset they used to fine-tune the pre-trained model is APCD. In one of the proposed experiments, the model was constrained to generate poetry from a specific meter. For evaluation, they used the BLUE score as well as human evaluation where they showed that this fine-tuning procedure outperformed all proposed approaches in the literature. They also showed another study with fake-generated poetry presented to subjects with limited poetic knowledge. They showed that the generated poetry was able to fool at least 61\% of the population.


\subsection{Poetry Identification from the Web}
\label{subsec:poetry_id_classificaiton}

\cite{almuhareb2013recognition} proposed a system to identify poetry from a text document. The proposed system relies extensively on the structural patterns of textual poetry. The system is evaluated on collected data from the web. The dataset has 23K lines with 161 classical poem instances. The method achieved an F-measure of 95\%. \cite{almuhareb2015recognition} extended their work by considering modern poetry that is different in style than classical poetry. The method is similar to the one with classical poetry in the sense that it focuses on the structural patterns of modern poetry. The method was evaluated on a dataset of 2,067 plain text documents containing 513 modern Arabic poems. The method achieves an accuracy of more than 99.81\%.
\cite{almuhareb2016arabic} developed a system for recalling Arabic poetry material from the web. The system consists of two main components, a classifier, and a distiller. The classifier classifies whether a page contains poetic material while the distiller absorbs the poetic material from the selected page. The system achieves a precision of 94\% on an initially selected 14 domains as a seed list.

\subsection{Miscellaneous Tasks}
\label{subsec:other_tasks}

\cite{khan2014using} applied the Arud meter system as a stenography tool. The idea is that the poem will be used as a cover message. Its binary representation is used to hide the secret message with the help of some special Arabic characters like diacritics. They compared their approach with other methods in the literature and they showed that their method outperforms others in the literature in the capacity score. \cite{abandah2020classifying} investigated the model architecture proposed by \cite{abandah2020accurate} which was designed for prose text to automatically diacritize Arabic poetry. They evaluated the model on an extended version of the dataset proposed by \cite{yousef2019learning}. They selected samples where the diacritization ratio is 0.5 or higher resulting in 368.6K verses. The results they showed are 6\% and 20\% for DER and WER respectively where it was 1.9\% and 5.1\% respectively for prose.
 
\section{Datasets}

The release of large Arabic poetry datasets did not happen until recently with the surge of deep learning. The first sufficiently large dataset published were, MetRec \cite{al2020metrec},  APCD \cite{yousef2019learning}, and APCD 2.0 \cite{abandah2020classifying}. MetRec is the smallest among the three of these datasets. It contains verses from the most frequent 14 meters of Arabic poetry with a total of 55.4K verses. APCD is a massive dataset compared to MetRec with more than 1.8M verses containing samples from all 16 meters. APCD was extended by \cite{abandah2020classifying} introducing APCD2.0. They added another class for prose to distinguish poetry from prose in their proposed classification model. Ashaar dataset extends APCD by adding more poetry while considering more sources. We also added a column for the poem theme which was not available in APCD. Table \ref{tab:apcd_ashaar_comparision} compares APCD with Ashaar. As can be seen from this table, Ashaar is almost an order of magnitude larger than APCD in terms of verses and poets. This plenty of poetic data along with poets is useful for many tasks concerning poetry generation such as language modeling and authorship attribution. It can be noted from the table that Ashaar is also larger in terms of diacritized verses. In this comparison, we considered verses where diacritics constitute more than 25\% of its characters. This is helpful for tasks that involve diacritics predictions.

\begin{table}[]
\centering
\begin{tabular}{lcc}
                     & \textbf{APCD}      & \textbf{Ashaar}    \\ \hline \hline
\ Poems             & -         & 254,630   \\ \hline
\ Poets             & 3,569     & \textbf{7,167}     \\ \hline
\ Verses             & 1,831,770 & \textbf{3,857,429} \\ \hline
\ Verses with meter   & 1,739,436 & \textbf{1,947,648} \\ \hline
\ Verses with theme  & -         & 1,757,639 \\ \hline
\ Verses with era    & 1,831,770 & \textbf{1,899,567} \\ \hline
\ Diacritized verses & 817,756   & \textbf{1,389,564} \\ \hline
\end{tabular}
\caption{Comparison between Ashaar dataset and APCD on different aspects.}
\label{tab:apcd_ashaar_comparision}
\end{table}

\begin{table*}[]
\begin{center}
\caption{Comparison between our model and \cite{abandah2020classifying}. We compare our models as a combination of dataset size and diacritics training. Also, we show the speed of running inference per 1024 batch size. }
\label{tab:meter_results}
\begin{tabular}{l|l|l|l|l}
 \textbf{Model/Metric} & \textbf{Diacritics}& \textbf{Training Corpus size} & \textbf{Prediction Time / 1024} & \textbf{Accuracy}  \\
 \hline
\hline
 \cite{abandah2020classifying} & \checkmark &1,493,086 & 388 ms & 96.18 \% \\
 Transformer Model & \checkmark & 806,062 & 84 ms & 95.51 \% \\
 Transformer Model & \checkmark & 1,460,255 & 84 ms & \textbf{96.24} \% \\
 \hline
 \cite{abandah2020classifying} &  & 1,493,086 & 388 ms & 94.18 \% \\
 Transformer Model &  & 806,062 & 84 ms & 93.90 \% \\
 Transformer Model  &  &  1,460,255 & 84 ms & \textbf{95.22} \% \\
\end{tabular}
\end{center}
\end{table*}

\section{Poetry Classification}
\label{sec:cls}
In this section, we mainly discuss three types of classifications which are era, theme, and meter classification. In each subsection, we illustrate the dataset used and the architecture for training. 

\subsection{Meter Classification}
\label{sec:meter}
As discussed in the literature, there are mainly 16 meters that govern how each poem should be constructed. In this subsection, we discuss our approach to generating a system that can predict a meter for a given poem. 

\paragraph{Preprocessing and Augmentation}
We first remove duplicates from the training corpus that exist in the testing corpus. Then for each verse in the training corpus, we split the two parts using a special symbol \texttt{\#}. We then remove all special characters except for the hashtag and diacritics. After that, we augment the corpus by randomly splitting each bait using \texttt{\#} then randomly swap the first and second parts. Also, to make the corpus more robust against partial diacritization at each step of training we randomly remove diacritics. We end up with 1,717,948 verses for training. We use a 15\% subset for validation. For testing, we use a dataset of size 362,798 verses.

\begin{figure}[ht]
    \centering
    \includegraphics{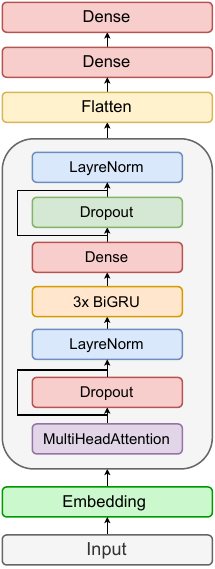}
    \caption{Model architecture for the meter classification model.}
    \label{fig:meter_model}
\end{figure}



\paragraph{Training and Results}
We use a transformer base model with multi-head attention. We start with an embedding of size 64. We use a transformer block with two dense layers at the end with ReLU activation functions. The transformer block contains, multi-head attention followed by dropout and layer normalization with a skip connection. We then add 3 Bidirectional GRU layers followed by one dense layer. The last block contains the same skip connections as in the previous block. In Figure \ref{fig:meter_model}, we show the main architecture of the model. We train the model for 15 epochs and we save the model that achieves the best validation accuracy. In Table \ref{tab:meter_results}, we compare the results of our transformer base model to the work of \cite{abandah2020classifying}. We mainly compare training on the smaller dataset and larger dataset with and without diacritics. Our base model trained on comparable corpus compared to the state of the art achieves better results with and without diacritics on the test set. Also, our models are 4 times faster in terms of inference when evaluated on a Tesla T4 machine with around 16GB of memory.

\subsection{Era Classification}
\begin{figure}[ht]
\includegraphics[width=7cm]{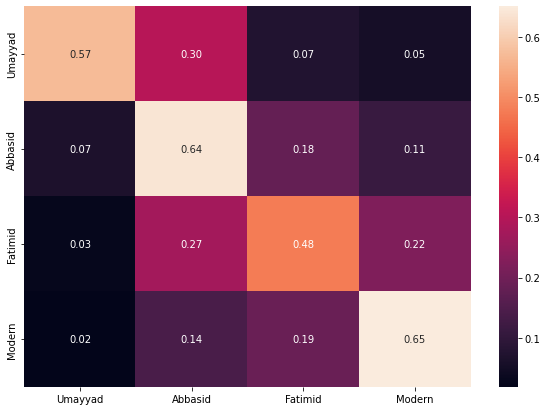}
\caption{Confusion matrix for era classification.}
\label{fig:era_results}
\end{figure}

We group the classes of poems into four main eras in Hijri corresponding to 1) before Islam - 132, 2) 132-232, 3) 232 - 784, and 4) 784-now. We use a max length of 64 verses for each poem. We then fix the max size of each class into 50,000 poems in order to avoid bias towards classes with many poems. For tokenization, we use a sentence-piece tokenizer and we create a model with a 10,000 vocabulary size with 128 max number of tokens for each poem. We train a model with 3 bidirectional layers and two dense layers with a dropout of size 30\% for 5 epochs with batch size 128. Figure \ref{fig:era_results}, shows the confusion matrix on the test set. We can notice that in general, we see confusion, especially in consecutive eras. 
\begin{table}[]
\centering
\caption{Time distribution of Arabic Poetry. The dates are in Hijri which is calculated using the Lunar calendar. }
\begin{tabular}{l|l}
\textbf{Era} & \textbf{Date (Hijri)}\\
\hline \hline
Umayyad & 041 - 132 \\ \hline
Abbasid & 132 - 232 \\ \hline
Al-Andalus & 113 - 172 \\ \hline 
Fatimid & 358 - 567 \\ \hline 
Ayyubid & 569 - 626 \\ \hline
Mamluk & 648 - 784 \\ \hline
Ottoman & 699 - 1200 \\ \hline
Modern & 1200 - now
\end{tabular}
\end{table}

\subsection{Theme Classification}
\begin{figure}[ht]
\includegraphics[width=7cm]{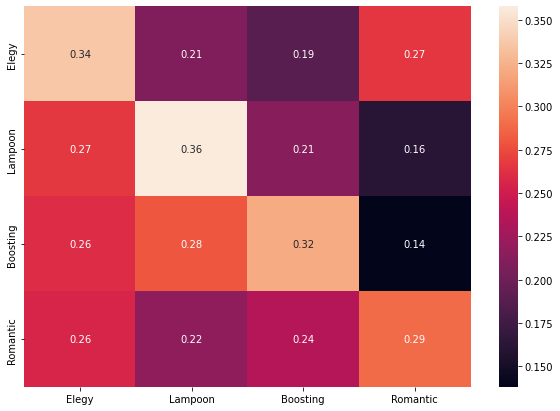}
\caption{Confusion matrix for theme classification.}
\label{fig:theme_results}
\end{figure}

We group the classes of poems into four main categories that are, elegy (sad) poems, lampoon (sarcasm) poems, boosting (praise) poems, and romantic poems. We use a similar training setup as in era classification. Figure \ref{fig:theme_results}, shows the confusion matrix on the test set. Generally, we observe that the model finds it much more difficult to predict the correct classes as compared to the era classification. We think the reason is the contamination of the dataset which might contain a lot of incorrect labels.

\begin{table*}[]
\begin{center}
\caption{Diacritization metrics on the test set of \textbf{Ashaar}.}
\label{tab:diac_results}
\begin{tabular}{l|l|l|l|l}
  \textbf{pre-training} & \textbf{DER} & \textbf{WER} & \textbf{DER*} & \textbf{WER*}  \\
 \hline
 
\hline
 Tashkeela  & 20.40 \% & 62.3 \% &   18.22 \%&  50.42 \% \\
 Ashaar   &  \textbf{14.03} \% & \textbf{47.97} \% &  \textbf{12.09} \% &  \textbf{36.09} \% \\
\end{tabular}
\end{center}
\end{table*}

\section{Poetry Diacritization}
\label{sec:diac}
In this section, we try to tackle the problem of diacritizing Arabic poetry. Usually, poetry contains many classical words and metaphors which makes assigning diacritics to sentences more challenging. 

\subsection{Training datasets}
We use the Tashkeela dataset for pre-training the model \cite{zerrouki2017tashkeela}. Since the dataset doesn't contain any splits, we utilize the splits suggested by \cite{fadel2019arabic} which contains 50k training, 2.5k validation, and 2.5k testing. For Ashaar, since there are many sentences that are not diacritized we filter by percentages of diacritics. If the verse has more than 5\% missing diacritics we discard it from training the model. We end up with 26,091 poems after also discarding short poems. We use 23,481, 1,305, and 1,305 for training, validation, and testing respectively. We utilize the word error rate (WER), diacritics error rate (DER), WER without case ending called WER*, and DER without case ending DER*.

\subsection{Results}
We use a 1-D convolution bank, highway network, and bidirectional GRUs from \cite{madhfar2020effective} as our main model for pre-training. We pre-train two models, one on Tashkeela and another on the diacritized version of Ashaar. We train each model for 10,000 steps and evaluate both on the test subset of Ashaar. In Table \ref{tab:diac_results}, we compare the two training strategies for diacritization. We observe that pre-training and then evaluating on Ashaar provide better results. 

\section{Predicting Arudi Style}
Each given meter has a closed set of tafeelat that represent how the meter should be constructed. For example, the Taweel meter has this sequence where 1 represents harakah and 0 represents a sukun: 

\begin{equation*}
\begin{aligned}
    &\color{teal} 11010 \, 1101010 \, 11010 \, 110110 \\
\end{aligned}
\end{equation*}

When a verse or hemistich is created it should follow one of the permissible representations. If the verse doesn't follow the meter, we can map it to the original sequence by \textcolor{blue}{addition}, \textcolor{red}{deletion}, or \textcolor{orange}{flipping}. As an example, the following sequence could be mapped to the previous sequence using that coloring scheme:

\begin{equation*}
\begin{aligned}
        &\color{teal}1101\color{blue}0\color{red}0 \, \color{teal}1101010 \, 1101\color{orange}1 \, \color{teal}110110
\end{aligned}
\end{equation*}

Using that representation, we can predict whether a given poem has any problems as the following. 

\begin{enumerate}
    \item We first created a dataset of all permissible changes of a given meter.
    \item For a given poem we diacritize it using the approach mentioned in section \ref{sec:diac}. We then map every harakah to 1 and sukun to 0. 
    \item Then we use our collected dataset to find the sequence with the largest cosine similarity match. We utilize the built-in function in python \texttt{SequenceMatcher} which gives a similarity score between input patterns. We can use our meter classification model to also reduce the cost of the search. 
\end{enumerate}

This makes our algorithm robust because even if the verse doesn't follow any given Tafeelah if the diacritized form is not accurate we will still predict the Tafeelah with high confidence. Furthermore, using our color coding representation we will be able to predict if a given character needs to be \textcolor{blue}{added}, \textcolor{red}{deleted} or \textcolor{orange}{flipped}.
In order to assess the ability of our system to predict correctly a given Arudi style, we created manually an independent test set containing 100 hemstitches. We use our system to predict the patterns and then compare the gold patterns to the output. Using that we get an average score of 93.41\% which indicates a high similarity score. We get 43\% with an exact match i.e. similarity score of 100\% which indicates a precise approach. 









\begin{table*}[!htp]
\begin{center}
\caption{Descriptions of the special tokens used in the tokenizer.}
\label{tab:tokens}
\begin{tabular}{l|l}
 \textbf{Functionality} & \textbf{Tokens}    \\ \hline \hline
 Poem separators & \texttt{["<|psep|>", "</|psep|>"]} \\ \hline
 Bait separators & \texttt{["<|bsep|>", "</|bsep|>"]} \\ \hline
 Verse separator & \texttt{["<|vsep|>"]} \\ \hline
 Themes & \texttt{["<|theme\_0|>",..., "<|theme\_17|>"]} \\ \hline
 Meters & \texttt{["<|meter\_0|>",..., "<|meter\_15|>"]} \\ \hline
 Unused tokens & \texttt{["<|res\_0|>",..., "<|res\_9|>"]} \\ \hline
 Other tokens & \texttt{["<|pad|>", "<|endoftext|>"]}
 
 \end{tabular}
\end{center}
\end{table*}

\section{Poetry Generation}
\label{gen}
In this section, we consider training a poetry model from scratch rather than fine-tuning. Our early experiments show that usually poetry doesn't work well with word pieces (see Appendix \ref{sec:appendix}) so instead we retrain the whole model on characters.

\subsection{Data Preparation}

\textbf{Representation} Our main objective is to train a model that can generate poetry that preserves the meter, theme, and structure of classical poetry. To do that, we introduce new types of tokens to the model as in Table \ref{tab:tokens}. 
Below we show a simple example of how to encode a given poem that contains two verses. We use an HTML-like prompting approach to be applied for a given input poem. Note that \texttt{n} is in the \texttt{range(0, 15)} and \texttt{k} in the range \texttt{range(0, 17)}. 

\warning{

\texttt{<|meter\_n|> qafyiah} 
\texttt{<|theme\_k|>}

\texttt{<|psep|>}

\tab \texttt{<|bsep|>}

\tab \tab \texttt{verse\_1<|vsep|>verse\_2}

\tab \texttt{</|bsep|>}

\texttt{</|psep|>}
}
    
Note that, for poems that don't have a meter we use our pre-trained meter classification to predict that \texttt{meter}. To make the prediction more robust, we use a majority vote over the poem to be more accurate. We filter out poems that don't match our meter classifier label. For the theme, we reserve a token for unknown \texttt{theme}.

\textbf{Data Cleaning and Filtration} We apply the following cleaning procedures for each poem
\begin{enumerate}
    \item We map characters using their Unicode representation. 
    \item Remove poems that don't have an even number of verses.
    \item Remove poems that have very small verses i.e number of characters less than 5. 
    \item Remove poems with meters that are not one of the 16 classes we have.
\end{enumerate}

We release our dataset pre-processed in that format in HuggingFace \footnote{\url{https://huggingface.co/datasets/arbml/Ashaar_dataset}}. 
\subsection{Training}
For this task, we don't remove any diacritics and we consider this as an approach to generate poetry with diacritics as well. Training a GPT-based model using BPE tokenization will be expensive because the frequency of word pieces will be much less, especially with partially diacritized text. So, we use a character-based tokenization approach. We train the model for 450,000 steps with batch size 16 and context size 512. The max vocabulary size is 121 which equals the number of characters plus diacritics in the corpus in addition to the reserved tokens in Table \ref{tab:tokens}. We use the default GPT-2 transformer-based architecture\footnote{\url{https://huggingface.co/docs/transformers/model_doc/gpt2}} with 10 layers. 

\begin{table*}[!htp]
    \centering
    \caption{Predicting conditional poetry based on the qafiyah, meter, and theme of the poem.}
    \label{tab:modern_samples}
    \begin{tabular}{p{6cm}|p{6cm}}
 \hline \hline
\begin{arabtex}
\textbf{قصيدة حزينه من بحر الكامل مع قافية بحرف الكاف}
يا صاحبيّ أفي تحمُّلِ ذاكا 

فمتى متى يُمْحى هواكا 

يا نومَ عينِى ما أطالَ تجلّدي

طيفُ الخيالِ يطولُ في هجراكا
         \end{arabtex}
         
         &
         
         \begin{arabtex}
\textbf{قصيدة غزل من بحر البسيط مع قافية بحرف الهاء}
ما لي أراك كأنّي مستهاماً 

بها افتخرتَ بلذاتي وأبلاها 

فلذةُ القلبِ في أحشاء مسكنها 

ولذة العيش في أنات أحلاها
         \end{arabtex}
         
         \\ \hline\hline
    
    \end{tabular}
    
\end{table*}

\begin{table*}[!htp]
\begin{center}
\caption{Comparison to State of the Art in rhythm evaluation results. We compare the tokenizer, number of layers Top-n accuracy where n is varied across 1, 3, and, 5.}
\label{tab:meter_following}
\begin{tabular}{p{4.5cm}|l|c|c|c|c}
 \textbf{Model} & \textbf{Tokenizer} & \textbf{Accuracy} & \textbf{Top-3 Accuracy}&
 \textbf{Top-5 Accuracy}
 \\
\hline \hline \cite{abboushi2023toward} & BPE & 56.70 & - & - \\
\hline 
\textbf{Ashaar} & Char & \textbf{64.40} &\textbf{69.40 \%} & \textbf{71.13} \\   

\end{tabular}
\end{center}
\end{table*}

\subsection{Evaluation}
Evaluating language models is a difficult task, let alone poetry generation which is a creative challenging task. For this purpose, we use a set of novel evaluation metrics, to evaluate the generative power of our pre-trained models. 

\begin{figure}[!htp]
\includegraphics[width=0.45\textwidth]{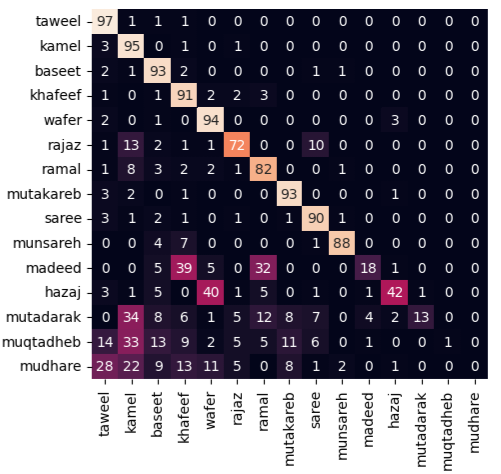}
\caption{Confusion matrix for rhythm evaluation.}
\label{fig:meter_following_results}
\end{figure}

\paragraph{Rhythm Evaluation}
In order to evaluate how much rhythm is encoded in our generated poetry we use meter classification for such a task. Given a generated poetry output we can evaluate how much the model can generate poetry that belongs to the same meter with high confidence. We use the same meter classification model that we created in section \ref{sec:meter}. Because we can not force the model to generate certain poetry, we use the model which gets a high accuracy to evaluate how much rhythm is able to generate. At each step, we generate 10 verses for the 15 meters used in \cite{abboushi2023toward}. We repeat the process 100 times for each meter resulting in 1,500 generated poems. Then, we pass the generated poems to our classification model to predict the meter. We use majority voting to decide if the poem meter is correct. For Top-3 and Top-5 accuracy, we predict correctly if one of the top 3 and 5 predicted meters contains the true meter. In Table \ref{tab:meter_following}, we show the results and compare them to the work done by \cite{abboushi2023toward}. Even though, our model is much smaller it, still achieves better results on the poem level. 
In Figure \ref{fig:meter_following_results}, we show the confusion matrix for meter classification on the generated poetry. We mostly observe that the more popular the meter the better results. Still, for 50\% of the meters, we achieve more than 90\%.

\paragraph{Zero-shot Analysis}
Zero-shot evaluation is used to evaluate how much pre-trained models can incorporate or generalize to new tasks without explicit pre-training or fine-tuning. The model was not pre-trained explicitly to predict diacritics for a given text in a supervised way. In Table \ref{tab:zero_shot}, we evaluate the correctness of our model in predicting diacritics. We evaluate the model against our pre-trained diacritization model in Section \ref{sec:diac}. We consider our model as the gold prediction. We pre-train an unconditional character-based model on Ashaar and evaluate its diacritization ability. We sample with different probabilities and evaluate the DER and WER metrics. We observe that the model is able to predict diacritics with at most a 50 \% error rate. 

\begin{table}[!htp]
\begin{center}
\caption{Zero shot evaluation on diacritics. We compare an unconditional pre-trained character-based GPT in zero-shot dicaritization abilities with different sampling rates.}
\label{tab:zero_shot}
\begin{tabular}{l|l|l}
 \textbf{Sampling Probability} & \textbf{WER} & \textbf{DER}  \\
 \hline
\hline
 sampling with 2 & 88.77 \% & 43.30 \% \\
 sampling with 3 & 88.57 \% & 44.62 \% \\
 sampling with 5 & 90.71 \% & 47.85 \%  \\
 sampling with 7 & 91.18 \% & 50.28 \% 
\end{tabular}
\end{center}
\end{table}

\section{Conclusion}
\label{conc}
To summarize, our paper introduces a system called \textit{Ashaar} capable of analyzing and generating conditional poetry. Additionally, we curate multiple datasets and assess their effectiveness in various tasks such as classification, diacritization, Arud prediction, and conditional poetry generation. Furthermore, we leverage this dataset to generate poetry and evaluate the performance of our character-based model in diacritization, where we observe a satisfactory level of proficiency. 

\section{Acknowledgement}
We would like to thank the colleague of Computing and Mathematics at KFUPM for providing the compute to train some of the models. Also, we would like to thank ML Collective for providing compute part which was used to train the generative models. We also would like to thank Omar Hammad, with whom we discuss some of the ideas during the early phases of the project. In addition, we would like to thank Kyle McDonald for providing the compute to pre-train some of the earlier models. 

\bibliography{anthology,custom}
\bibliographystyle{acl_natbib}

\appendix

\section{Modern Poetry Generation}
\label{sec:appendix}
In this section, we discuss our experiments of pre-training on modern poetry. 

\subsection{Pre-training Dataset}

This is a large dataset collected from around 10 Arabic newspapers \cite{el20161}. Arabic newspapers are written in Modern Standard Arabic (MSA) which really meets our needs. One more advantage of such newspapers is their diversity. Topics are written in many domains, such as sports, politics, economy, etc. The total size of the text is about 15GB with 1.5 billion words. Working on such large datasets needs special techniques even in reading and extracting some useful statistics. We read it in chunks in parallel settings to speed up. We mainly use a GPT-2 architecture with context size 512 with 12 heads and 12 layers. In order to fit the model on a V100 with a reasonable number of batch sizes we trained a model with the following parameters:  We reduced the context size (which refers to the number of tokens or history to train in parallel) to half which is a reasonable compromise because usually poems are not long. For training, we used the transformer-lm which is a Pytorch implementation that allows training in multiple GPUs. We trained on a 4 x V100 machine with a 200 GB HDD drive, 32 virtual CPUs, and 120 GB memory. The speed-up was around 2.6x compared to one GPU. However, in each epoch, the speed up was reduced which might have been caused by some memory leak. The 120 GB memory was necessary to tokenize the 14GB dataset all at once. Training using SentencePiece tokenizer. Training on the full corpus for around 10 epochs took around 20 days.

\begin{table*}[!htp]
    \centering
    \begin{tabular}{p{5cm}|p{5cm}}
    \hline
    \hline
         \begin{arabtex}
         يا أمة الإسلام هبي من سباتك ...
 و اخلدي من سباتك 
 ...
 فلقد غفوت قرون و اضطرب الوجود 
 ...
 منذ أجيال و شعوب تداولها 
 ...
 سكرة الأحلام حين الشعور 
 ...
 وفتون الأعالي من تراه يعودها 
 ...
 ذكرى أرق و دنيا دفينة لا ترجع
         \end{arabtex}
         &
         \begin{arabtex}
     وقالوا : إني سأمضي
     ...
     فكنتين معي 
     ...
     لكي تحسي بقايانا 
     ...
     وراكضين على الدروب  
     ...
     بين المروج الحالمات 
     ...
     بعطر الهوى 
     ...
     ونسائم الربيع 
     ...
     حيث الهوى يقضم شفاهنا 
     ...
     ويأكل من رحيق الوداد 
     ...
     فكرة حائر

         \end{arabtex}
         
         \\ \hline
         \begin{arabtex}
         والأمر كل الأمر أنك غائبه ...
 وأنا حضرت لأجملك بالغياب 
 ...
 وقبلك كنت أعرف أنني لا أحسن سوى الكلام 
 ...
 فكيف لك الحضور  ؟  
 ...
 لست الغياب 

         \end{arabtex}
         & 
         \begin{arabtex}
         الآن يا حبيبتي ... هل تعلمين 
         ...
 ما معنى دمي الذي ينساب مبلولا في الفراغ 
 ...
 وكآبات النخيل على الأفق الحزين 
 ...
 وكأني أغزل الآن أنشودة 
 سوداء 
 ...
 وأستعير 
 ...
 من ذكرى الرحيل إلها للمساء 

         \end{arabtex}
\\
\hline
\hline
    \end{tabular}
    \caption{Predicting samples in free form.}
    \label{tab:modern_samples}
\end{table*}

\subsection{Fine-tuning}

There is a big shift in terms of context writing style and vocabulary from the poetry of old traditional Arabic and the Modern Standard recent one. This will cause undesirable results as the model is trained on a Modern Standard Arabic text! We tried to limit or eliminate the old poetry to the best of our capabilities.
In terms of the datasets collected, we collected poetry from three famous Arabic poetry websites: Aldiwan, Abudhabi poetry encyclopedia, and Adab. 
Aldiwan and Abu Dhabi encyclopedia sources hold, kind of, similar poetry in terms of structure, and categorization methodology. On the other hand, Adab is more comprehensive and contains more recent poetry that is close to MSA in terms of vocabulary and context and also more prose poetry than the other two repositories. It is the best fit to be used in the transfer learning task. Aldiwan and Abudhabi encyclopedia sources hold, kind of, similar poetry in terms of structure, and categorization methodology. On the other hand, Adab is more comprehensive and contains more recent poetry that is close to MSA in terms of vocabulary and context and also more prose poetry than the other two repositories. It is the best fit to be used in the transfer learning task.

 We fine-tuned the pre-trained model on a poetry dataset. We experimented with different approaches to see what is the effect on the generated poetry.  First, we trained on a poetry dataset that is based on meters extracted from Aldiwan. However, we realized that the model was not able to preserve the meter. Basically, it jumped between different meters. To see the effect of training on a certain meter, we extracted all the poems that belong to a certain meter which is Taweel. However, we realized something interesting. The generated poetry was somehow meaningless because the model tried to pick specific words to preserve the chosen meter. GPT-2 is a subword model but to generate a poem from a certain meter, it has to have some knowledge about characters. Moreover, there are 16 meters and training on the highest class (Taweel) reduced the size of the dataset to only around 15 MB. Arabic is a morphologically rich language and in order to capture language understanding we need a larger dataset. Moreover, most of them were old and contained many words that are not used anymore in modern literature. This created a vocabulary shift between pre-training and fine-tuning. 
In order to overcome these we only extracted modern poems from Adab. We created an algorithm to clean the dataset and remove short poems as well as rhythmic poems. We ended up with around a 26 MB dataset. We applied the same process for segmentation and tokenization as we did for our pre-training. In addition, we added special characters to recognize the end of poems “\#” and “\&” for the end of verses. Training without these special characters mostly caused incoherent results. We also did a lot of cleaning in order to increase the quality of generated poems. We made sure to normalize the characters (Some characters had different Unicode so we made sure that they are mapped to the same set of characters), remove digits (some poems had digits indicating different parts of a given poem), remove special characters (there was a lot of, metadata and diacritics. We realized that the quality of the dataset affected a lot the generated poems in some way or another. We did a lot of back and forth by fine-tuning, inspecting the output, cleaning then fine-tuning again. 

Then we fine-tuned our pre-trained model for around 200 epochs with the same parameters. This time we fine-tuned the model on Google Colab with a single V100 GPU for around three days. To see the effect of training longer we analyzed the results after each 50 iteration. To see if the model memorized some poems we randomly extracted some poems and ran the model for inspection. We realized that the model learned a lot of variations in the generated poems. We applied some post-processing approaches to increase the readiness of the generated poems. We first segmented, replaced “\&” and “\#” with new lines, and resolved some issues with FARASA which created some unneeded characters.  For inference, we append the special character “\#” to the prefix indicating the start of a new poem. We use the top 3 predictions for randomly predicting the next token. We tested with larger numbers but it resulted in some bad output. In Table \ref{tab:modern_samples}, we show a sample of predicted modern poetry. 

\end{document}